\documentclass{article}

\PassOptionsToPackage{numbers, compress}{natbib}

\usepackage[final]{nips_2018}

\usepackage[utf8]{inputenc} 
\usepackage[T1]{fontenc}    
\usepackage{hyperref}       
\usepackage{url}            
\usepackage{booktabs}       
\usepackage{amsfonts}       
\usepackage{nicefrac}       
\usepackage{microtype}      

\usepackage{graphicx}
\usepackage{arydshln}
\usepackage{caption}
\usepackage{subcaption}
\usepackage{amsmath}
\usepackage{amssymb}
\usepackage{multirow}
\usepackage{arydshln}
\usepackage{booktabs}
\usepackage{xcolor}
\usepackage{wrapfig}

\title{Adversarial Domain Adaptation for Stance Detection}

\author{
Brian Xu, Mitra Mohtarami, James Glass\\
Computer Science and Artificial Intelligence Laboratory\\
Massachusetts Institute of Technology\\
Cambridge, MA, USA \\
  \texttt{\{bwxu, mitram, glass\}@mit.edu} \\
}

\begin{document}

\maketitle

\begin{abstract}
This paper studies the problem of stance detection which aims to predict the perspective (or stance) of a given document with respect to a given claim. Stance detection is a major component of automated fact checking. As annotating stances in different domains is a tedious and costly task, automatic methods based on machine learning are viable alternatives. In this paper, we focus on adversarial domain adaptation for stance detection where we assume there exists sufficient labeled data in the source domain and limited labeled data in the target domain. Extensive experiments on publicly available datasets show the effectiveness of our domain adaption model in transferring knowledge for accurate stance detection across domains.
\end{abstract}

\section{Introduction}\label{introduction}

With the rise of social media and microblogs, there has been an increasing awareness of the negative influence of fake news and how it can unfairly influence public opinion on various events and policies~\cite{mihaylov-georgiev-nakov:2015:CoNLL,ACL2016:trolls,Vosoughi1146}. In order to counteract these effects, various organizations are now performing manual fact checking on suspicious claims. However, manual fact checking can't feasibly keep up with the sheer volume of fake claims. A fact-checking process for a given claim is a challenging multi-step process, that typically involves the following steps~\cite{vlachos2014fact}:
(\emph{i})~retrieving potentially relevant documents as evidence for the claim~\cite{AAAIFactChecking2018,R17-1046},
(\emph{ii})~predicting the stance of each document with respect to the claim~\cite{mitra2018memory,baly2018integrating},
(\emph{iii})~estimating the trustworthiness of the documents (e.g. in the Web context, the site of a Web document could be an important indicator of its trustworthiness), and finally
(\emph{iv})~making a decision based on the aggregation of (\emph{ii}) and (\emph{iii}) for all documents from (\emph{i})~\cite{AAAIFactChecking2018}.

In this paper, we focus on the second step of the fact-checking process which is the \textit{stance detection} task. Stance detection aims to automatically determine the perspective (or stance) of a document to a claim as \textit{agree}, \textit{disagree}, \textit{discuss}, or \textit{unrelated}. Since there is not enough data with annotated stance labels for many domains, machine learning algorithms can easily produce poor to mediocre performance. One potential approach to alleviate the lack of labeled data is to supplement the data sources with data from other domains. However, it is not a straightforward process as each data source generally has its own unique distributions and nuances. In this paper, we tackle this problem using a transfer learning technique, namely adversarial domain adaptation, to effectively use labeled data from a source domain to improve stance detection performance on a target domain which has limited data. Our contributions can be summarized as follows:
\begin{itemize}
\item We are the first to apply adversarial domain adaptation to the problem of stance detection across different sources.
\item Our model outperforms the best baseline on a publicly-available benchmark dataset, Fake New Challenge~\footnote{Available at \url{www.fakenewschallenge.org}}. 
\end{itemize}

\section{Method}\label{method}

\begin{figure}
\centering
\includegraphics[width=99mm,scale=1]{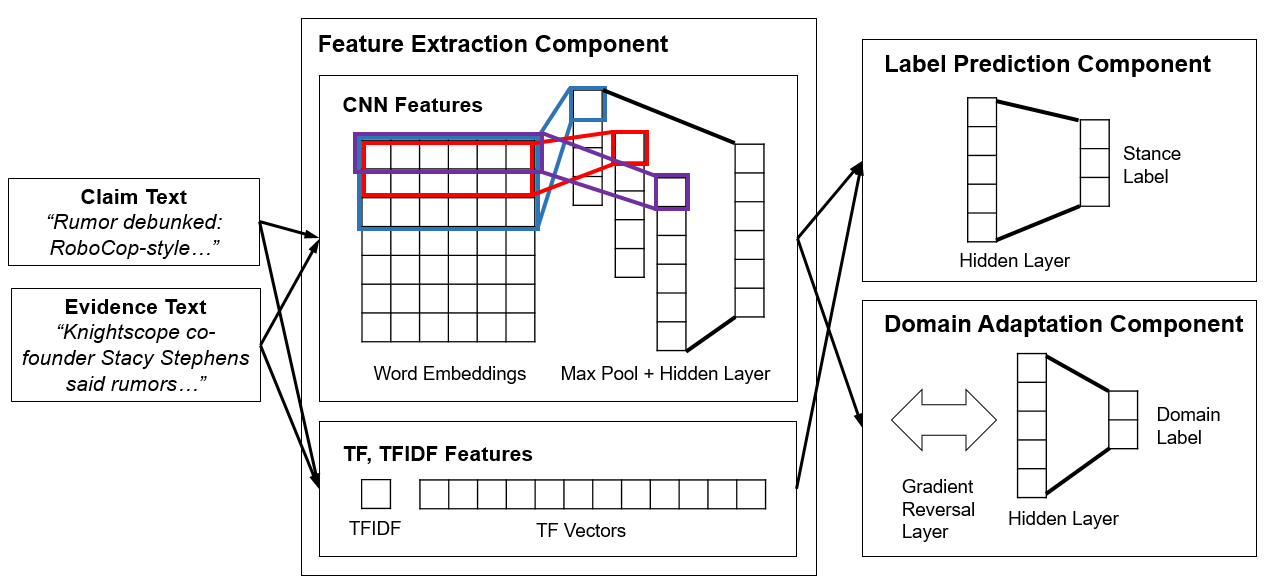}
\caption{The architecture of our model with domain adaptation component for stance detection.}
\label{fig:model_arch}
\vspace{-5pt}
\end{figure}

Previously-proposed approaches for stance detection generally contain two components~\cite{Talos,hanselowski2017athene,DBLP:journals/corr/RiedelASR17}: a feature extraction component followed by a class label prediction component. In this paper, we present a model for stance detection that augments the traditional models with a third component: a domain adaptation component. Our domain adaptation component uses adversarial learning~\cite{2014arXiv1409.7495G} to encourage the feature extraction component to select common--rather than domain-specific--features when input data is from multiple different domains. This allows the model to better leverage \textit{source} domain data for better prediction on data from the \textit{target} domain. The general architecture of our model is shown in Figure~\ref{fig:model_arch}.
As illustrated, the inputs are first given to the ``\textit{Feature Extraction Component}'' to compute their features and representations. These features are then passed to the ``\textit{Label Prediction Component}'' and then to the ``\textit{Domain Adaptation Component}.'' In the model, while the both latter components try to minimize their own losses, the feature extraction component attempts to maximize a domain classification loss to encourage better mixture of examples from different domains. The components of the model are described in detail below.

\textbf{Feature Extraction Component:}
This component takes the input claim $c$ and document $d$ and converts them to their semantic representations and features. To do this, in this component, we use bag-of-words (BOW)---e.g., TF and TF.IDF-weighted features---and cosine similarity between $c$ and $d$ features. We select these BOW features as these features are useful to filter the documents with `unrelated' stance labels as shown in section~\ref{model-schemes}. 
Furthermore, we also use a Convolutional Neural Network (CNN) approach~\cite{DBLP:journals/corr/Kim14f} for learning representations of claims and documents. We use a CNN because it can capture $n$-grams and long-range dependencies~\cite{DBLP:journals/corr/YuHBP14}, and can extract discriminative word sequences that are common in the training instances~\cite{severyn2015learning}. These traits make CNNs useful for dealing with long documents~\cite{Mohtarami:2016:SemEval}.

\textbf{Label Prediction Component:}
This component uses a Multi-Layer Perceptron (MLP) with a fully-connected hidden layer followed by a \textit{softmax} layer which employs \textit{cross entropy} loss as the cost function. This component will predict stance labels as \textit{agree}, \textit{disagree}, \textit{discuss}, or \textit{unrelated} for a given set of claim and document features.

\textbf{Domain Adaptation Component:}
We introduce a domain classifier which includes a MLP followed by a \textit{softmax} layer. Given a set of features for a claim-document pair, the domain classifier predicts which domain the features originated from. 
The domain classifier is an adversary because the model---specifically, the feature extraction component---attempts to maximize the domain classifier loss, while the domain classifier attempts to minimize it. This is because a high domain classifier loss implies that the domain classifier is unable to accurately discern whether a set of features belongs to a \textit{source} or \textit{target} domain. This implies that the features extracted from the input examples are common to both the \textit{source} and \textit{target} domains as we desire. 
To achieve this adversarial effect, the features from the feature extraction component are passed to a \textit{gradient reversal} layer before being passed to the domain classifier. The gradient reversal layer is a simple identity transform during forward propagation and multiplies the gradient by a negative constant (the gradient reversal constant) during backpropagation~\cite{2014arXiv1409.7495G}. By adding the gradient reversal layer, the desired training behavior can be achieved through normal model training.

\section{Experiments and Evaluations}\label{results}

\subsection{Datasets}
We use the Fake News Challenge (FNC) dataset$^1$ as \textit{target} data. This data is collected from a variety of sources such as rumour sites, e.g. \url{snopes.com}, and Twitter accounts such as \textit{@Hoaxalizer}. It contains around $50$K claim-document pairs as training data with an imbalanced distribution over stance labels: $73$\% (\textit{unrelated}), $18$\% (\textit{discuss}), $7.3$\%(\textit{agree}), $1.7$\%(\textit{disagree}). We note that there is a lack of labeled data especially for \textit{agree} and \textit{disagree} stance labels in FNC. Furthermore, we use the Fact Extraction and VERification (FEVER) dataset~\cite{FEVER} as \textit{source} data. This dataset is collected from Wikipedia and contains around $145$K claim-document pairs as training data with imbalanced distribution over stance labels: $55$\% (\textit{supported}), $21$\% (\textit{refuted}), $24$\%(\textit{Not Enough Information (NEI)}). We discard the examples with NEI labels as there is not any documents assigned to the claims in those examples, and correspond the \textit{supported} and \textit{refuted} labels in FEVER to the \textit{agree} and \textit{disagree} labels in FNC respectively.

\subsection{Evaluation Metrics}
We use the following evaluation metrics: 
\textbf{Macro-F1}: The average of the $F_1$ score for each class. 
\textbf{Accuracy}: The number of correctly classified examples divided by their total number of examples.
\textbf{Weighted-Accuracy}: This metric is presented by FNC~\footnote{The scorer is available at \url{www.fakenewschallenge.org}} which is a two-level scoring scheme. It gives $0.25$ weight to the correctly predicted examples as \textit{related} or \textit{unrelated}. It further gives $0.75$ weights to the correctly predicted \textit{related} examples as \textit{agree}, \textit{disagree}, or \textit{discuss}.

\subsection{Model Parameters and Training Procedure}
For our CNN, we use $300$-dimensional \texttt{Word2Vec}~\cite{DBLP:journals/corr/abs-1301-3781,DBLP:journals/corr/MikolovSCCD13,DBLP:conf/naacl/MikolovYZ13} word embeddings trained on \texttt{GoogleNews} dataset\footnote{\url{https://code.google.com/p/word2vec}}, and $128$ feature maps with filter width \{$2$, $3$, $4$\}. We set maximum word lengths of $50$ and $500$ for claims and documents respectively; these values are greater than the length for most claims and documents in the target train data. For the BOW model, we keep the hyper-parameters and features the same as the baseline model~\cite{DBLP:journals/corr/RiedelASR17}. 
Our models are trained using the \texttt{Adam} optimizer, and $20\%$ of the training data is set aside as validation data. In the models with a domain adaptation (DA) component, equal amounts of both \textit{source} and \textit{target} data are randomly selected at each epoch during training. Finally, we fine-tune all the hyper-parameters of our models on validation data which contains equal amounts of \textit{source} and \textit{target} data.

\subsection{Baselines}\label{baselines}
We compare our domain adaptation (DA) model to the following baselines: (a)~\textit{Gradient Boosting}, which is the Fake News Challenge baseline that trains a gradient boosting classifier using hand-crafted features reflecting polarity, refute, similarity and overlap between documents and claims; (b)~\textit{TALOS}~\cite{Talos}, which was ranked first at FNC. It uses hand-crafted features as well as 
weighted-average between gradient-boosted decision trees (TALOS-Tree) and a deep convolutional neural network (TALOS-DNN); (c)~\textit{UCL}~\cite{DBLP:journals/corr/RiedelASR17}, which was ranked third at FNC. This model trains a softmax layer using $n$-gram features (e.g., TF and TF.IDF). We compare with this model as our BOW model is similar to this model and uses the same features. 

\subsection{Our Models}\label{model-schemes}

We present different variations of our models where each uses a subset of components/features shown in Figure~\ref{fig:model_arch}. These variations help us to conduct ablation analyses on this information.
The baseline and our models are trained to predict stance labels on target data; \{\textit{agree}, \textit{disagree}, \textit{discuss}, \textit{unrelated}\}. We further apply a two-level \textit{hierarchy} prediction scheme in our models. That is, a model with a \textit{hierarchical} scheme is first trained to predict two stance labels as \{\textit{unrelated} or \textit{related}\}, and then the examples predicted as \textit{related} are only given to the model to predict the labels \{\textit{agree}, \textit{disagree}, \textit{discuss}\}. For the first step of the \textit{hierarchy} scheme, we use the BOW model which achieves a high F$_1$ performance of $97.7$\% for \textit{unrelated} and $93.9$\% for \textit{related} labels.

\begin{table*}[t]
\centering
\caption{Results on the FNC test data. BOW, CNN and DA refer to our model when it uses bag-of-words features, convolutional features, and domain adaptation, respectively. When DA is present, square brackets indicate which features are passed to the domain adaptation component. The \textit{hierarchy} in parentheses refers to our model with two-level prediction scheme as explained in section~\ref{model-schemes}. We show the results of the models based on the smallest loss for validation set across $5$ independent runs.}
\scalebox{0.70}{
\begin{tabular}{lccccc}
  \toprule
	\bf Model & \bf Train Data & \bf Weighted-Acc. & \bf Acc. & \bf Macro-F$_1$ & \bf F$_1$ ({\it agree, disagree, discuss, unrelated})\\
  \midrule
  1. \ \ \ \ \ \multirow{1}*
    {Gradient Boosting} & FNC & 75.2 & 85.4 & 45.7 & 14.8 / 2.0 / 69.5 / 96.5 \\
  2. \ \ \ \ \ \multirow{1}*
    {TALOS {\it \small (\#1st in FNC)}} & FNC & 82.0 & 89.1 & 57.8 & 53.8 / 3.6 / 76.0 / 97.9 \\
  3. \ \ \ \ \ \multirow{1}*
    {TALOS-DNN} & FNC & 60.8 & 66.5 & 41.8 & 27.6 / 9.3 / 47.4 / 82.7 \\
  4. \ \ \ \ \ \multirow{1}*
    {TALOS-Tree} & FNC & \underline{83.1} & \underline{89.5} & 56.8 & 53.4 / 0.2 / \underline{76.3} / \underline{98.4} \\
  5. \ \ \ \ \ \multirow{1}*
    {UCL {\it \small (\#3rd in FNC)}} & FNC & 81.7 & 88.5 & 57.9 & 47.9 / 11.4 / 74.7 / 97.6 \\
  \midrule
  6. \ \ \ \ \ \multirow{1}*
    {BOW} & FNC & 81.1 & 88.6 & 56.0 & 49.2 / 2.5 / 74.8 / 97.6 \\
  7. \ \ \ \ \ \multirow{1}*
    {CNN} & FNC & 40.8 & 71.3 & 23.3 & 0.3 / 0.0 / 10.0 / 83.0 \\
  8. \ \ \ \ \ \multirow{1}*
    {BOW + CNN} & FNC & 74.9 & 86.8 & 52.2 & 41.4 / 0.0 / 72.1 / 95.2 \\
  9. \ \ \ \ \ \multirow{1}*
    {BOW (hierarchy)} & FNC & 80.7 & 88.5 & 57.3 & 49.5 / 7.5 / 74.3 / 97.7 \\
  10. \ \ \ \multirow{1}*
    {CNN (hierarchy)} & FNC & 79.9 & 87.9 & 56.0 & 54.9 / 0.2 / 71.1 / 97.7 \\
  11. \ \ \ \multirow{1}*
    {BOW + CNN (hierarchy)} & FNC & 80.3 & 88.2 & 56.5 & \underline{56.0} / 0.0 / 72.1 / 97.7 \\
  \midrule
    12. \ \ \ \multirow{1}*
      {BOW} & FNC, FEVER & 78.5 & 86.4 & 56.3 & 48.8 / 9.8 / 69.4 / 97.1 \\
    13. \ \ \ \multirow{1}*
      {[BOW + DA]} & FNC, FEVER & 72.9 & 81.5 & 48.6 & 44.5 / 2.0 / 51.7 / 96.1 \\
    14. \ \ \ \multirow{1}*
      {BOW (hierarchy)} & FNC, FEVER & 78.8 & 87.3 & 57.2 & 51.7 / 10.2 / 69.1 / 97.7 \\
    15. \ \ \ \multirow{1}*
      {[BOW + DA] (hierarchy)} & FNC, FEVER & 78.5 & 87.1 & 56.4 & 51.6 / 8.3 / 68.0 / 97.7 \\
  \midrule
  16. \ \ \ \multirow{1}*
    {CNN} & FNC, FEVER & 41.3 & 64.3 & 27.3 & 17.4 / 2.0 / 11.0 / 78.8 \\
  17. \ \ \ \multirow{1}*
    {[CNN + DA]} & FNC, FEVER & 39.0 & 64.3 & 24.0 & 13.5 / 0.2 / 3.3 / 78.9 \\
  18. \ \ \ \multirow{1}*
    {CNN (hierarchy)} & FNC, FEVER & 79.0 & 87.4 & 56.6 & 51.9 / 7.5 / 69.3 / 97.7 \\
  19. \ \ \ \multirow{1}*
    {[CNN + DA] (hierarchy)} & FNC, FEVER & 79.1 & 87.7 & 57.9 & 51.2 / 11.4 / 71.3 / 97.7 \\
  \midrule
  20. \ \ \ \multirow{1}*
    {BOW + CNN} & FNC, FEVER & 71.7 & 84.5 & 51.5 & 44.6 / 5.6 / 60.2 / 95.6 \\
  21. \ \ \ \multirow{1}*
    {BOW + [CNN + DA]} & FNC, FEVER & 71.9 & 84.6 & 51.4 & 44.9 / 4.4 / 60.6 / 95.6 \\
  22. \ \ \ \multirow{1}*
    {BOW + CNN (hierarchy)} & FNC, FEVER & 79.6 & 87.8 & 56.6 & 53.1 / 5.1 / 70.6 / 97.7 \\
  23. \ \ \ \multirow{1}*
    {BOW + [CNN + DA] (hierarchy)} & FNC, FEVER & \bf{80.3} & \bf{88.2} & \underline{\bf{60.0}} & \underline{\bf{54.6}} / \underline{\bf{15.1}} / \bf{72.6} / \bf{97.7} \\
\bottomrule
\end{tabular}}
\label{tab:results}
\vspace{-10pt}
\end{table*}

\subsection{Performance Analysis}
Table~\ref{tab:results} shows the results of different models for the target test data, i.e., FNC test data. Lines $1$-$5$ show the results for baseline models explained in section~\ref{baselines}. As the results show, they weakly perform for \textit{agree} and \textit{disagree} stances due to the small size of labeled data; only $1.7$\% and $7.3$\% of the FNC data has \textit{disagree} and \textit{agree} labels. 
Lines $6$-$23$ show the results of different configurations of our model as explained in section~\ref{model-schemes}. Lines $6$-$8$ show the results for BOW, CNN and their combinations respectively; the results of these models with the \textit{hierarchy} scheme are shown in lines $9$-$11$. The results show that using the \textit{hierarchy} scheme can help models to perform better, especially for CNN model where its result improves from $23.3$\% F$_1$ (line $7$) to $56$\% (line $10$). However, these models don't improve the baseline results, except for the combination of BOW and CNN models with the \textit{hierarchy} scheme (see line $11$). This model achieves the best result for \textit{agree}, while it poorly performs on \textit{disagree}. These results show that a promising solution is to use domain adaptation to deal with the limited data size as we show below.

While the above results (lines $1$-$11$) are obtained when only the FNC train data is used, lines $12$-$23$ show the results of different configuration of our model when we use additional training data from a different domain (i.e., FEVER) in addition to the FNC data. Lines $12$-$15$ show the results for the BOW model with and without the \textit{hierarchy} scheme and domain adaptation (DA) component. The results show that domain adaptation does not improve the results for BOW model. 
We repeat these experiments with CNN model (lines $16$-$19$), and the CNN model with \textit{hierarchy} scheme can significantly perform better when using the DA component. Given these results, we pass only convolutional features to the DA component in lines $20$-$23$ which show the results for our model when using a combination of BOW and CNN model with and without the \textit{hierarchy} scheme and domain adaptation. The model that combines BOW and CNN model with the \textit{hierarchy} scheme and domain adaptation achieves the best F$_1$ performance, especially for \textit{disagree} and \textit{agree} classes---because the \textit{source} data only contains the corresponding \textit{disagree} and \textit{agree} labels. 
In summary, the results show:
\begin{itemize}
\vspace{-3pt}
\item The \textit{hierarchy} scheme can help our models to perform better across all the metrics.
\vspace{-3pt}
\item Our best model is the combination of BOW, CNN+DA, and the hierarchy scheme. It outperforms the baselines on F$_1$, especially on the most important classes: \textit{disagree}, \textit{agree}.
\vspace{-3pt}
\item The \textit{source} FEVER data can improve the performance of our model for target FNC data through adversarial domain adaptation, when uses CNN model with \textit{hierarchy} scheme (see lines $18$-$19$ in Table~\ref{tab:results}) or BOW+CNN model (see lines $20$-$23$ in Table~\ref{tab:results}).
\end{itemize}

\subsection{Training Loss Trend}\label{discussion}

\begin{wrapfigure}{r}{0.53\linewidth}
\vspace{-10pt}
\centering
  \includegraphics[width=70mm,scale=1]{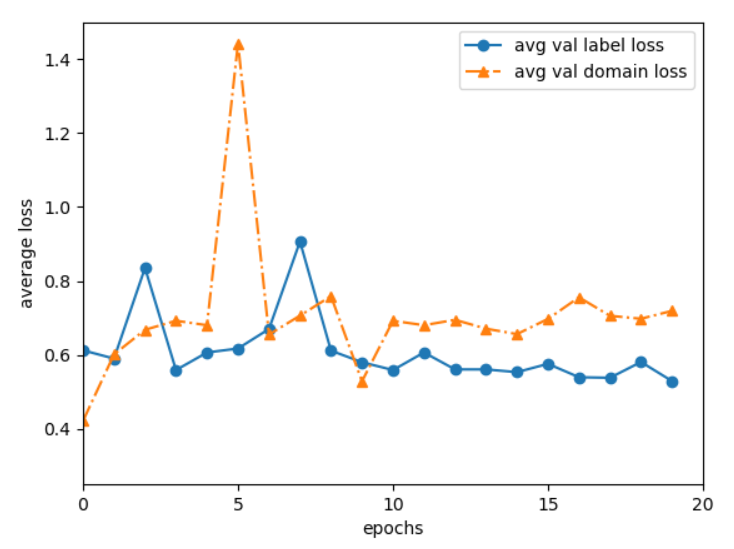}
  \caption{The classification (or label) and domain adaptation losses on validation data across training epochs for our best model; BOW + CNN + DA with \textit{hierarchy} scheme.}
  \label{fig:val_loss}
\vspace{-27pt}
\end{wrapfigure}

Figure~\ref{fig:val_loss} shows the classification and domain adaptation losses across epochs on validation data for our best model: BOW + CNN + DA model with the \textit{hierarchy} scheme. 
As shown in the figure, both losses are unstable during the early epochs of training. This is because of learning and domain adaptation rates, which couples a high learning rate with a ramp up in the domain adaption rate.
Then, after around $10$ epochs, the classification and domain adaptation losses get more stable and behave as expected; the classification loss slowly decreases as the label prediction component in the model attempts to minimize its loss, while the domain adaptation loss slowly increases as the model attempts to maximize the domain loss so that it cannot distinguish between its \textit{source} and \textit{target} examples as we explained in section~\ref{method}.

\section{Conclusion}\label{conclusion}

We present a model that uses adversarial domain adaptation for the task of stance detection. Our experiments show the effectiveness of our model in transferring knowledge among stance datasets, from the FEVER dataset to the FNC dataset. Our model outperforms the state-of-the-art approaches for this task and obtains $60$\% F$_1$ on the FNC target data.
For future work, we plan to apply our best model with adversarial domain adaptation to other combinations of datasets collected for stance detection. Furthermore, we plan to use the datasets collected for a similar task, e.g., the Stanford Natural Language Inference (SNLI) data~\cite{snli:emnlp2015} to investigate the utility of our model in transfer learning between inference and stance detection tasks.


\bibliography{references}
\bibliographystyle{plainnat}
\end{document}